\pdfoutput=1

\documentclass[11pt]{article}

\usepackage[final]{acl}

\usepackage{times}
\usepackage{latexsym}

\usepackage[T1]{fontenc}

\usepackage[utf8]{inputenc}

\usepackage{microtype}

\usepackage{inconsolata}

\usepackage{graphicx}
\usepackage{amssymb}
\usepackage{enumitem}
\usepackage{amsmath}
\usepackage{autobreak}
\usepackage{multirow}
\usepackage{svg}
\usepackage{booktabs}
\usepackage{subcaption}
\usepackage{tcolorbox}
\usepackage{listings}
\newcommand{\cut}[1]{}
\title{Cognition Chain for Explainable Psychological Stress Detection \\on Social Media}

\author{Xin Wang$^{1}$, Boyan Gao$^{1}$, Yi Dai$^{2}$, Lei Cao$^{3}$, Liang Zhao$^{4}$, Yibo Yang$^{1}$, David Clifton$^{1,5}$\thanks{Corresponding author.}\\
  $^1$Department of Engineering Science, University of Oxford \\
  $^2$Department of Computer Science and Technology, Tsinghua University \\ $^3$Faculty of Psychology, Beijing Normal University \\ 
  $^4$School of Information Management, Wuhan University \\
$^5$Oxford Suzhou Centre for Advanced Research \\
}

\begin{document}
\maketitle
\begingroup
\renewcommand{\thefootnote}{}
\footnotetext{Emails: \{xin.wang, david.clifton\}@eng.ox.ac.uk}
\endgroup
\begin{abstract}
Stress is a pervasive global health issue that can lead to severe mental health problems, including anxiety, depression, and suicide. Early detection offers timely intervention and prevention of stress-related disorders. The current early detection models perform "black box" inference suffering from limited explainability and trust which blocks the real-world clinical application. Thanks to the generative properties introduced by the Large Language Models (LLMs), the decision and the prediction from such models are semi-interpretable through the corresponding description. However, the existing LLMs are mostly trained for general purposes without the guidance of psychological cognitive theory. 
 To this end, we first highlight the importance of prior theory with the observation of performance boosted by the chain-of-thoughts tailored for stress detection. This method termed Cognition Chain explicates the generation of stress through a step-by-step cognitive perspective based on cognitive appraisal theory with a progress pipeline: Stimulus → Evaluation → Reaction → Stress State, guiding LLMs to provide comprehensive reasoning explanations.
We further study the benefits brought by the proposed Cognition Chain format by utilising it as a synthetic dataset generation template for LLMs instruction-tuning and introduce CogInstruct, an instruction-tuning dataset for cognitive stress detection. This dataset is developed using a three-stage self-reflective annotation pipeline that enables LLMs to autonomously generate and refine instructional data. By instruction-tuning Llama3 with CogInstruct, we develop CogLLM, an explainable stress detection model. Evaluations demonstrate that CogLLM achieves outstanding performance while enhancing explainability. Our work contributes a novel approach to stress detection by integrating cognitive theories into LLM reasoning processes, offering a promising direction for future explainable AI research.
\end{abstract}

\section{Introduction}

\begin{figure}[t]
\centering 
\includegraphics[width=1\linewidth]{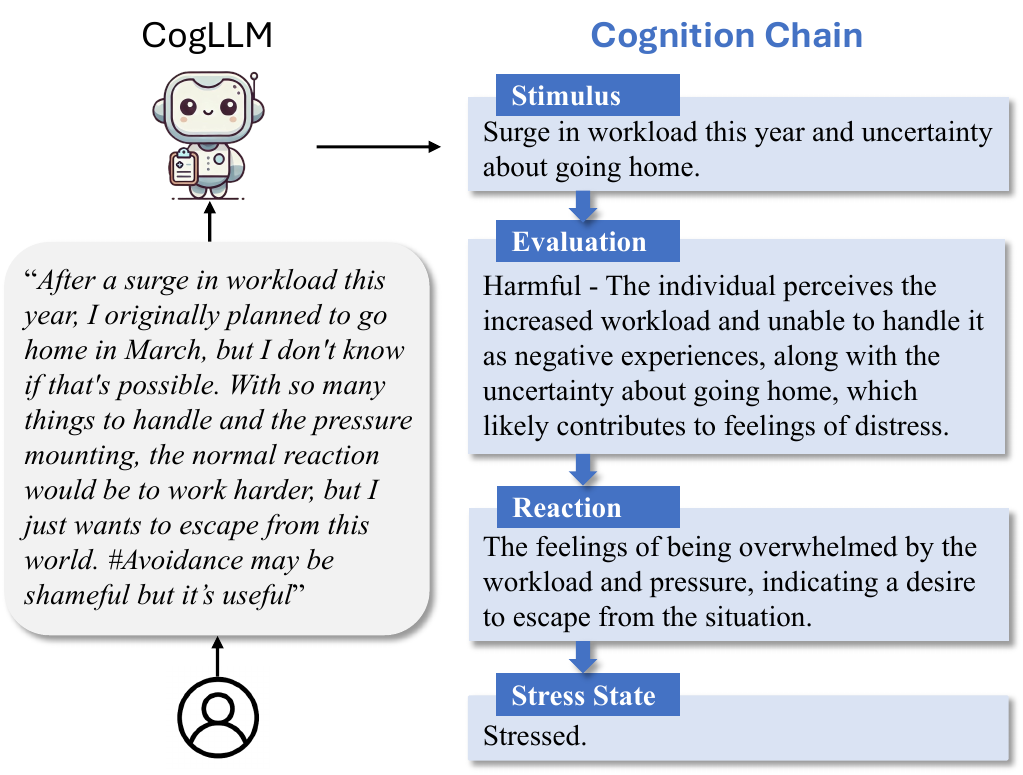}
\caption{An example of our Cognition Chain. The left part is an individual's post. The right part is our Cognition Chain.}
\label{fig:introduction}
\end{figure}

Stress is a pervasive issue, causing severe mental health issues such as anxiety, depression, self-harm, and suicide, which significantly impair individuals' health worldwide~\cite{harmofstress}. Fortunately, timely interventions can prevent the escalation of stress-related disorders through early stress detection, which improves overall well-being and reduces long-term health risks. 

Differing from traditional counseling \cite{goodman1984going} and questionnaires \cite{holmes1967social} based stress
detection methods, daily expression reflects a person's mental state in real-time. Building on this, natural language-based approaches, particularly those analyzing social media content \cite{guntuku2019understanding}, leverage platforms like Twitter, Reddit, and Weibo, where individuals increasingly share emotions, thoughts, and experiences. These methods enable real-time analysis and provide insights into societal trends.

Despite advancements in stress detection algorithms, lacking explainability in model predictions remains a significant challenge, rendering the results difficult to trust and interpret. 
Most existing stress detection studies~\citep{guntuku2019understanding,turcan2021emotion,wang2023contrastive} approach stress detection as a binary classification task, simply outputting 0 or 1 to indicate stress states. These models often operate as "black boxes", providing little to no explanation about how conclusions about an individual's stress states are drawn.
As a result, these unexplained outcomes may lead to user mistrust and hinder the adoption of such technologies in clinical settings. Therefore, integrating explainable methods into stress detection models is crucial for improving reliability, building trust, and supporting more effective decision-making.

Thanks to the generative properties introduced by the Large Language Models (LLMs), the prediction from such models is semi-explainable through the corresponding description.
Recently, a few works \cite{alghamdi2024less} have designed direct prompt to instruct LLMs to infer stress states. Due to the generative nature, LLMs will also generate some explanations for their predictions. However, the direct prompting method still have limitations. First, the generation process lacks guidance from psychological cognitive theories, leading to inadequate or unreasonable explanations of stress generation. Furthermore, since LLMs generate tokens sequentially, where each token depends on the previously generated ones, uncomprehensive or incorrect explanations can propagate errors and degrade the detection performance.

Cognitive appraisal theory~\cite{lazarus1991emotion,arnold1960emotion} explains individuals' emotional generation from a cognitive perspective. 
The theory suggests that the generation of human emotion can be summarized as `stimulus-appraisal-emotion'. For example, imagine receiving an unexpected job offer. The job offer is the stimulus. You appraise this event by considering factors like how it aligns with your career goals, the job location, salary, and potential challenges. If the appraisal leads you to see the offer as a great opportunity that advances your career, you may feel excited and happy. Conversely, if you perceive the job as unsuitable or threatening to your current work-life balance, you might feel anxious or stressed.

These inspire us to integrate cognitive appraisal theory into the Chain-of-Thought (CoT) prompt~\cite{wei2022chain} to achieve comprehensive reasoning explanations for stress detection. Specifically, we propose a \textbf{Cognition Chain}, defined as \textit{Stimulus $\rightarrow$ Evaluation $\rightarrow$ Reaction $\rightarrow$ Stress State}, to steer LLMs to explain the generation of stress step by step from a cognitive perspective. As shown in Figure \ref{fig:introduction}, the process begins with a stimulus, which undergoes cognitive evaluation as the individual assesses its implications. This evaluation shapes the subsequent emotional response and ultimately leads to a stress state. To operationalize this approach, we design a prompt template for LLMs to implement our Cognition Chain.

To train a model capable of generating effective Cognition Chain for explainable stress detection, we propose an instruction tuning dataset, \textbf{CogInstruct}, which is composed of cognition chain data. 
Specifically, we introduce a three-stage self-reflection pipeline. 
The pipeline begins by automatically generating Cognition Chain data using our prompt template and GPT-4o, followed by correcting erroneous data through self-reflection. Finally, any remaining incorrect data is corrected through answer-reflection. Beyond the automatic processes, we manually annotate a portion of the data and train a classifier to further eliminate low-quality data. 
Through the above process, we obtain CogInstruct.
Subsequently, we choose the Llama3, an open-source large language model as our base model. By instruction-tuning the Llama3 model on CogInstruct, we obtain the \textbf{CogLLM} model. 
The experimental results demonstrate that CogLLM not only achieves outstanding detection performance but also provides comprehensive explanations about stress generation.

We highlight the following contributions\footnote{The data and code will be release on publication.}:
\begin{itemize}
\item We propose the \textit{Cognition Chain}, a method to explain the generation of stress step by step from a cognitive perspective. We design a Cognition Chain prompt template to guide the reasoning processes of large language models. 
\item We construct \textit{CogInstruct}, an instruction-tuning dataset containing Cognition Chain data for explainable stress detection. Moreover, we propose a three-stage self-reflective annotation pipeline that enables LLMs to autonomously curate instruction datasets.  
\item We present \textit{CogLLM}, an instruction-tuned LLama3 on the Construct dataset. Experiment results demonstrate that CogLLM achieves significant performance improvements while enhancing explainability.
\end{itemize}

\section{Related Work}
\subsection{Stress Detection on Social Media}
These stress detection on social media studies aim to determine whether an individual is stressed or not~\cite{xue2016analysis,saha2017modeling,lin2017detecting}. 
Thelwall proposed TensiStrength~\cite{thelwall2017tensistrength}, a system for detecting stress and relaxation levels in social media text, using lexical and machine learning approaches.
Pillai et al.~\cite{gopalakrishna2018detection} developed an improved method by incorporating word sense disambiguation, enhancing the performance of the TensiStrength algorithm.
Guntuku et al.~\cite{guntuku2019understanding} developed domain adaptation models to predict stress using Facebook and Twitter data, with potential applications for monitoring stress at individual and county levels.
Wang et al. ~\cite{wang2020leverage} introduced a three-level personalized stress detection framework, which integrates individual's group characteristics and personality traits.
Turcan et al.~\cite{turcan2021emotion} developed emotion-infused models for psychological stress detection, using multi-task learning and fine-tuning with emotion detection tasks.
ContrasRL~\cite{wang2023contrastive} proposed contrastive learning tasks to enhance stress representation learning for social media-based stress detection.
More recently, Alghamdi et al.~\cite{alghamdi2024less} utilized large language models (LLMs) for stress detection by designing a simple prompt to tell GPT-3.5~\cite{ye2023comprehensive} summary and answer stressed or not.

However, these methods typically output the category of stress states but rarely provide comprehensive reasoning explanations, which is crucial for making the results more convincing and reliable in real-world applications. To address this issue, we propose Cognition Chain to guide large models in generating step by step reasoning processes. Additionally, we constructed the first instruction tuning dataset, CogInstruct, which includes cognitive reasoning processes, and developed CogLLM, the first large language model designed for explainable stress detection from a cognitive perspective.

\subsection{Chain-of-Thought Prompting}
Chain-of-Thought (CoT) prompting has gained significant attention in recent years as a technique to enhance the reasoning capabilities of large language models ~\cite{chu2024navigate}. 
Wei et al~\cite{wei2022chain} were the first to propose chain-of-thought prompting, where a few chain of thought demonstrations are provided as references in prompting. 
Subsequently, numerous studies have continued to explore ways to improve CoT prompting methods, such as: Zero-shot CoT~\cite{kojima2022large}, PAL~\cite{gao2023pal}, Auto-CoT~\cite{zhangautomatic}, AutoMate-CoT~\cite{shum2023automatic}, and BoostedPrompt~\cite{pitis2023boosted}. 
Zero-shot CoT~\cite{kojima2022large} is a method that essentially involves appending the phrase `Let's think step by step' to the original prompt.
Moreover, CoT has been used in some domains such as dialogue and medical systems. 
For example, Chae et al.~\cite{Chae23} proposed decomposing commonsense reasoning into sequential steps, generating rationale to enhance Dialogue CoT reasoning. 
Wang et al.~\cite{hongruwang23} proposed two dialogue CoTs that enable advanced reasoning and planning based on user statuses.
Lievin et al.~\cite{lievin2024can} directly adopted Zero-shot CoT to generate accurate responses to medical-related questions.
Mao et al.~\cite{miao2024chain} utilized and analyzed CoT application in answer to nephrology-related questions.
Hagan et al.~\cite{o2023accuracy} used Zero-shot CoT to improve the accuracy and appropriateness of ChatGPT responses on skin cancer Information.

Inspired by the above works, we design an improved Chain-of-Thought (CoT) prompting method for explainable stress detection. Our key innovation is integrating cognitive appraisal theory to propose the Cognition Chain that can generate comprehensive reasoning process.

\section{Methodology}
In this section, we present Cognition Chain, beginning with a brief overview of Chain-of-Thought reasoning and Cognitive Appraisal Theory. We then formalize the proposed reasoning process, following the \textit{Stimulus $\rightarrow$ Evaluation $\rightarrow$ Reaction $\rightarrow$ Stress State} framework, integrating these two complementary approaches. To generalize this reasoning process, we construct a prompt template, 
which is subsequently used to generate CogInstruct data for model instruction-tuning.
Finally, we tune CogLLM base on CogInstruct.
\subsection{Background for Chain-of-Thought}

We provide a brief background on standard chain-of-thought (CoT) reasoning~\cite{wei2022chain}. We define the notations as follows: $\mathcal{Q}$ denotes the question; $\mathcal{T}$ denotes the prompt; and $\mathcal{F}$ denotes the final answer. The CoT prompt $\mathcal{P} = { I, (x_1, e_1, y_1), \dots, (x_n, e_n, y_n) }$ consists of an instruction $I$ and a few examples, each comprising a question $x_i$, a rationale $e_i$, and an answer $y_i$. In chain-of-thought reasoning, the model generates step-by-step reasoning steps $\mathcal{G}$ before producing the final answer $\mathcal{F}$, as shown in the equation below: \begin{equation} p(\mathcal{F}, \mathcal{G} \mid \mathcal{P}, \mathcal{Q}) = p(\mathcal{G} \mid \mathcal{P}, \mathcal{Q}) \cdot p(\mathcal{F} \mid \mathcal{P}, \mathcal{Q}, \mathcal{G}) \end{equation}

\subsection{Background for Cognitive Appraisal Theory}
We improve chain-of-thought reasoning for stress detection drawing inspiration from the cognitive appraisal theory~\cite{lazarus1991emotion,arnold1960emotion}, which explains how emotions are generated through a specific mental process. In essence, this process can be encapsulated in the sequence: `\textit{stimulus-appraisal-emotion}'. It begins when an individual perceives or pays attention to particular stimuli in their environment or to specific events. These stimuli prompt the person to engage in internal evaluations or interpretations, as beneficial, harmful, or irrelevant. Based on this appraisal, a corresponding emotional response is triggered, shaping how the person feels and reacts to the situation. 
For example, an oncoming interview can be a stimulus, the individual’s appraisal can be beneficial or harmful, if negative, the stimulus may form a bad emotional response like anxiety.

\subsection{Cognition Chain Creation}
To achieve reliable stress detection,  
we propose a step-by-step \textbf{Cognition Chain}, given the insights of human cognitive process when forming stress~\cite{lazarus1991emotion,arnold1960emotion}.
The reasoning process can be summarized as \textit{Stimulus $\rightarrow$ Evaluation $\rightarrow$ Reaction $\rightarrow$ Stress State}. The specific meaning of each element in the chain is as follows:
\begin{itemize}[parsep=1pt]
    \item \textbf{Stimulus} $\mathcal{S}$ refers to any potential trigger that initiates an individual’s emotional cognitive process. It can be external, such as an event or object, or internal, such as a thought or memory. 
    \item \textbf{Evaluation} $\mathcal{E}$ refers to an individual’s personal interpretation, assessment, and internal reaction to a stimulus, shaped by their personality, experiences, beliefs, and expectations. The evaluation of the stimulus can result in three outcomes: "beneficial," "harmful," or "irrelevant" to the individual.
    \item \textbf{Reaction} $\mathcal{R}$ presents an individual's reaction or state and the corresponding emotions due to stimulus and evaluation. 
    \item \textbf{Stress State} $\mathcal{A}$ summarises whether the individual is stressed based on the previous inference results.
\end{itemize}
Given a large language model, the generation process of our Cognition Chain is as follows:
\cut{
\begin{equation}
\begin{aligned}
p(\mathcal{A}, \mathcal{S}, \mathcal{E}, \mathcal{R} \mid \mathcal{T}, \mathcal{C}) &= p(\mathcal{S} \mid \mathcal{T}, \mathcal{C}) \\
&\quad \cdot p(\mathcal{E} \mid \mathcal{T}, \mathcal{C}, \mathcal{S}) \\
&\quad \cdot p(\mathcal{R} \mid \mathcal{T}, \mathcal{C}, \mathcal{S}, \mathcal{E})  \\
&\quad \cdot p(\mathcal{A} \mid \mathcal{T}, \mathcal{C}, \mathcal{S}, \mathcal{E}, \mathcal{R})
\end{aligned}
\end{equation}
}
\begin{align}
p(\mathcal{A}, \mathcal{S}, \mathcal{E}, \mathcal{R} \mid \mathcal{T}, \mathcal{C}) &= p(\mathcal{S} \mid \mathcal{T}, \mathcal{C}) \\
&\quad \cdot p(\mathcal{E} \mid \mathcal{T}, \mathcal{C}, \mathcal{S}) \nonumber\\
&\quad \cdot p(\mathcal{R} \mid \mathcal{T}, \mathcal{C}, \mathcal{S}, \mathcal{E}) \nonumber  \\
&\quad \cdot p(\mathcal{A} \mid \mathcal{T}, \mathcal{C}, \mathcal{S}, \mathcal{E}, \mathcal{R}) \nonumber
\end{align}

where $\mathcal{C}$ denotes the individual's post content. $\mathcal{T}$ is the prompt template we propose for implementing our Cognition Chain. $\mathcal{S}$, $\mathcal{E}$, $\mathcal{R}$, and $\mathcal{A}$ represent the sequential steps.

\begin{figure}[t]
\centering 
\includegraphics[width=1\linewidth]{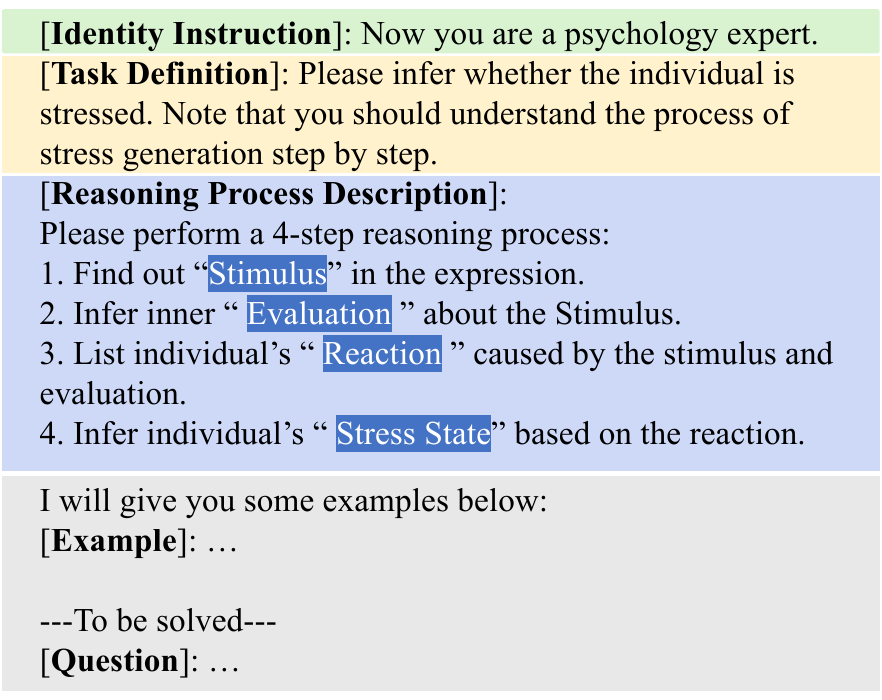}
\caption{Illustration of our proposed prompt template for the Reasoning Chain. The detailed prompt template is presented in Appendix.}
\label{fig:prompt template}
\end{figure}

\begin{figure*}[t]
\centering 
\includegraphics[width=1\linewidth]{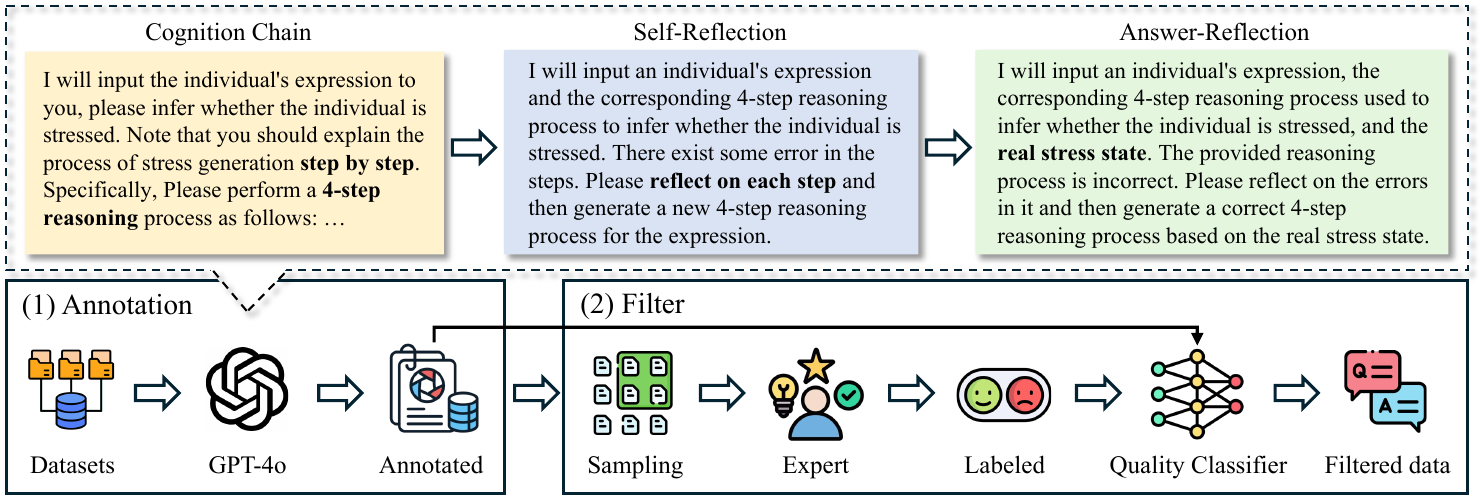}
\caption{Illustration of the constuction process for our CogInstruct.}
\label{fig:CogInstruct}
\end{figure*}

\subsection{Cognition Chain Prompt Template Design}

We designed a prompt template to guide LLMs in thinking step by step, forming what we refer to as our Cognition Chain. As illustrated in Figure \ref{fig:prompt template}, our prompt template consists of five main parts.
\begin{itemize}[parsep=1pt,label={-},leftmargin=*]
    \item \textbf{Identity Instruction}: This part is to prompt LLMs act as a psychological expert. 
    \item\textbf{Task Definition}: This part contains definition of the explainable stress detection task. Specifically, tell the LLMs what will be input, what to do, and what to output.
    \item \textbf{Reasoning Process Description}: This part describes the meaning of each element in the reasoning chain and the specific requirements of the 4-steps reasoning steps. We guide the model to explicitly list "stimulus", "evaluation", "response" and "stress state" as items to clarify the reasoning process.
    \item \textbf{Example}: This part provides a few manually annotated samples as references to help the LLMs better understand our requirements. Specifically, each example includes an individual's expression and the corresponding reasoning process.
    \item \textbf{Question}: This part presents the testing sample to be solved.  
\end{itemize}

\subsection{CogInstruct Dataset Construction}
We aim to build an instruction-tuning dataset containing Cognition Chains for training an explainable stress detection model. However, existing public stress detection datasets lack cognition chain data. Manually annotating a large amount of cognition chain data is undoubtedly costly because analyzing and writing cognition chains is very time-consuming. Furthermore, the required qualifications for labelers are high, usually necessitating psychology experts. Therefore, we introduce an AI-assisted expert approach. Specifically, we first use a three-stage self-reflection pipeline based on GPT-4 to obtain initial annotated data. Then, experts manually verify a small portion of the sampled data and use the verification results to train an annotation quality classifier. We use this quality classifier to eliminate low-quality samples from the initial annotated data, retaining only those of good quality for our final dataset.

\subsubsection{Self-Reflection Annotation}
We employ a powerful large language model (GPT-4o) to assist us in constructing the CogInstruct dataset. GPT-4o generates high-quality rationales when the derived answer is correct, as these rationales have been shown to effectively guide the model toward our expected outcomes. Unfortunately, it may also produce incorrect answers accompanied by misleading reasoning steps. LLM has been proven to have ability of self-correction~\cite{zelikman2022star}. Therefore, to generate and refine reasoning steps, we propose a three-stage self-reflection annotation pipeline comprising Cognition Chain Prompt, Self-reflection, and Answer-reflection stages.

We use raw data from two public datasets, dreaddit~\cite{turcan2019dreaddit} and WBSD~\cite{wang2023contrastive}. Merging their training sets yields a total of 5,295 samples. Each sample consists of a post and its label. We then annotate the cognition chains for each sample.  

1) \textbf{Cognition Chain Prompt Stage}: In this stage, we use GPT-4o and the Cognition Chain prompt template, as illustrated in Figure \ref{fig:prompt template}, to generate responses for each post. After filtering out the incorrect responses, we are left with 3,960 correct ones. The remaining incorrect responses, along with their corresponding posts, are then passed on to stage two.
\begin{figure*}[t]
  \centering
\begin{subfigure}[t]{0.24\linewidth}
\centering
\includegraphics[width=1\textwidth]{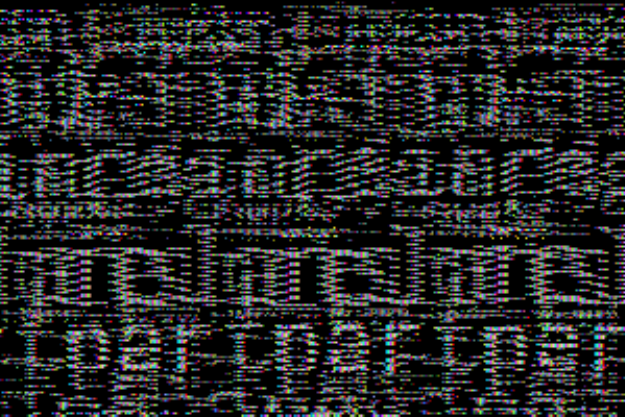}
\caption{Stimulus}
\end{subfigure}
\begin{subfigure}[t]{0.24\linewidth}
\centering
\includegraphics[width=1\textwidth]{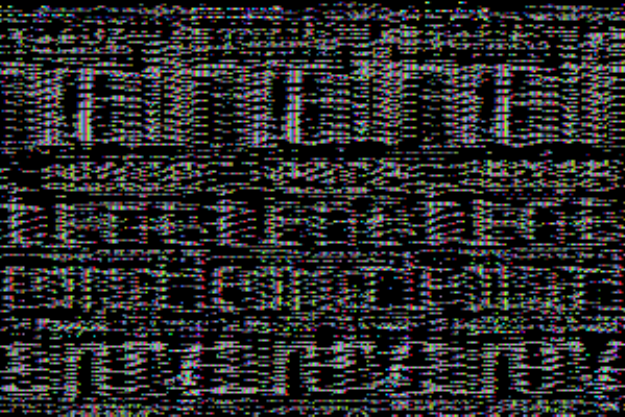}
\caption{Evaluation}
\end{subfigure}
\begin{subfigure}[t]{0.24\linewidth}
\centering
\includegraphics[width=1\textwidth]{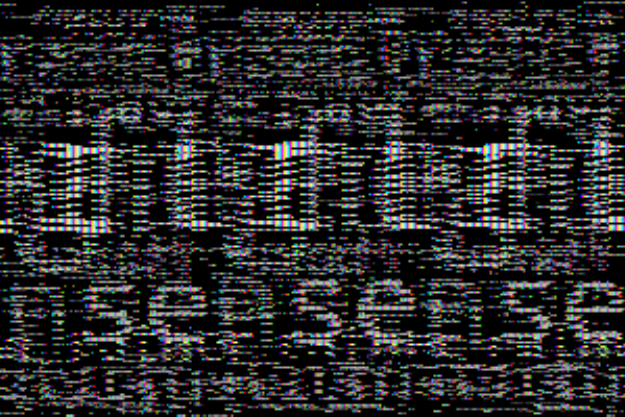}
\caption{Reaction}
\end{subfigure}
\begin{subfigure}[t]{0.24\linewidth}
\centering
\includegraphics[width=1\textwidth]{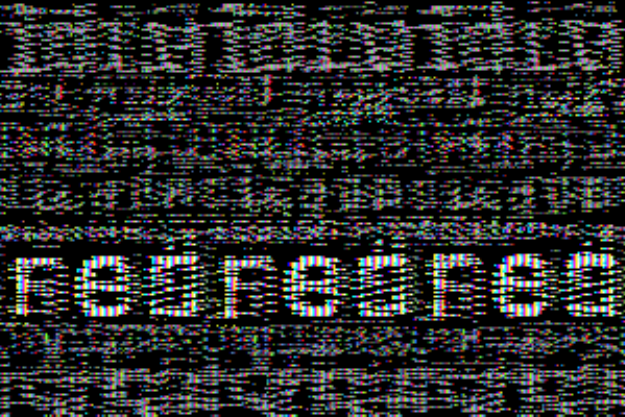}
\caption{Stress State}
\end{subfigure}
  \caption{The word cloud of each Cognition Chain step in our CongInstruct dataset}
  \label{fig:wordcloud}
\end{figure*}

2) \textbf{Self-Reflection Stage}: In this stage, we prompt GPT-4o to reflect on the samples with incorrect answers from the first stage. We inform GPT-4o that errors exist but do not provide the correct answers, encouraging it to review and analyze its previously generated intermediate steps. The newly generated responses are then filtered again, and we obtain 210 correct responses; the incorrect ones proceed to stage three.

3) \textbf{Answer-Reflection Stage}: In this stage, we prompt GPT-4o to review the samples with incorrect answers from the second stage. This time, we directly provide the correct answers, allowing GPT-4o to reflect and refine its understanding based on the final outcomes. We finally obtain 1,395 correct responses for the remaining posts. The entire process is illustrated in Figure \ref{fig:CogInstruct}. The detailed prompts for each stage are provided in the Appendix.

through the three-stage self-reflection process, we automatically added cognition chains to the original set of 5,295 samples. These augmented samples will be further filtered.

\subsubsection{Human Filter}
Although the previous process generates cognition chain reasoning steps, erroneous ones could be introduced due to two main reasons: 1) Even if the LLM generates the correct answer, the intermediate reasoning steps may still be flawed~\cite{lanham2023measuring}. 2) Despite specifying the output format, the model may occasionally produce extraneous information that deviates from the expected output.

We randomly selected a subset of 531 samples annotated by GPT-4o for manual verification. A group of professional psychology experts evaluated the quality of the automatically generated cognition chains by rating 388 samples as qualified and the remaining 143 as unqualified.

To improve the quality of our dataset, we developed a quality classifier using expert-annotated samples and RoBERTa~\cite{liu2019roberta}. Specifically, we labeled qualified samples as positive and unqualified samples as negative to train a binary classifier. This classifier determines whether the cognition chain quality of a sample meets the required standard. It was used to filter qualified samples into our CogInstruct dataset. After classification, we obtained 4,282 qualified samples, which were then converted into the instruction-tuning format. Experimental results in Appendix \ref{appdenix: filter} show that the model performs better on the filtered dataset.

As a result, our CogInstruct dataset consisted of 4,282 instructions. Table \ref{tab:statistic_CogInstruct} provides the dataset statistics, while Figure \ref{fig:wordcloud} presents the word clouds for each step in the Cognition Chain.

\subsection{CogLLM Tuning}
We follow the standard Supervised Instruction-Tuning (SIT) protocol to optimize pre-trained models using the CogInstruct dataset. Specifically, given a model parameterized by $\theta$ and input-answer pairs $\langle (\mathcal{T}, \mathcal{C}); (\mathcal{A}, \mathcal{S}, \mathcal{E}, \mathcal{R}) \rangle$, we minimize the losses generated under the next-token prediction paradigm, while masking out those associated with $\mathcal{T}$ and $\mathcal{C}$. The process can be expressed as:
\begin{align}
    \min_{\theta}\,\, & \mathcal{L}_\text{SFT}  ( (\mathcal{T},\mathcal{C}); (\mathcal{A},  \mathcal{S},\mathcal{E},\mathcal{R}))  \\ 
    & = -\log \text{p}_\theta(\mathcal{A},  \mathcal{S},\mathcal{E},\mathcal{R})|\mathcal{T},\mathcal{C}).\nonumber
\end{align}
Without losing generality, we choose Llama-3-8B-Instruct~\cite{touvron2023llama} as the pre-trained model regarding both the performance and computational cost.

\begin{table}[t]
\centering
\small
\begin{tabular}{lrr}
\toprule
\multirow{2}{*}{Post} & Num. of  Posts                     & 4,282 \\
                          & Avg. Length  per Post             & 68.7 \\ \midrule
\multirow{4}{*}{\begin{tabular}[c]{@{}l@{}}CogChain\\ (Avg. Length)\end{tabular}}      & Stimulus ($\mathcal{S}$)          & 14.8   \\
                          & Evaluation ($\mathcal{E}$)     & 21.3  \\
                          & Reaction ($\mathcal{R}$) & 25.6  \\
                          & Stress State ($\mathcal{A}$)      & 12.9  \\ \bottomrule
\end{tabular}
\caption{Statistics of our CogInstruct dataset.}
\label{tab:statistic_CogInstruct}
\end{table}

\section{Experiment}

\subsection{Experiment Details}
To prepare the dataset for tuning the large language model (LLM), we converted it into the Alpaca format. All experiments during the instruction-tuning process were conducted using the Huggingface Transformers library. We employed LoRA instruction-tuning~\cite{hulora} for training. The learning rate is set to 1.0e-4 and the training epoch is set to 3.
We provide 2 examples in our Cog-Prompt, which are also provided as reference for other baseline methods.
For all experiments, we report the average performance of 5 times running. Our algorithm is implemented with
PyTorch 2.4.0 on a Linux server RTX 4090 GPUs.

\subsection{Datasets}
We conduct experiments using two human-annotated stress detection datasets.
1) Dreddit~\cite{turcan2019dreaddit}. This dataset consists of 3,563 Reddit posts, each manually labelled as stressful or non-stressed. The original data is split into a training set and a testing set, with 2,838 and 715 posts, respectively. We further divide the data into training, validation, and testing sets with proportions of 2,838/358/357.
2) WBSD~\cite{wang2023contrastive}. This dataset contains 3,098 Weibo posts, which have same label space as Dreddit and are annotated by multiple psychology experts. The training, validation, and testing sets consist of 2,475, 307, and 316 posts, respectively.

\begin{table}[t]
\small
\setlength\tabcolsep{3.5pt}
\begin{tabular}{@{}llcccc@{}}
\toprule
\multirow{2}{*}{\textbf{Model}} & \multirow{2}{*}{\textbf{Method}} & \multicolumn{2}{c}{\textbf{Dreaddit}} & \multicolumn{2}{c}{\textbf{WBSD}} \\ \cmidrule(l){3-6} 
                              &                                  & Acc          & F1-score          & Acc        & F1-score        \\ \midrule
\multirow{3}{*}{GPT-3.5}      & Prompt                           & 70.14             & 77.42             & 75.67            & 79.92           \\ & CoT                           & 70.50             & 77.65             & 76.84            & 80.86           \\
                              & \textbf{CogChain}                         & \textbf{73.25}    & \textbf{79.84}    & \textbf{79.25}   & \textbf{83.48}  \\ \midrule
\multirow{3}{*}{Gemini-1.5}   & Prompt                           & 72.28             & 77.51             & 77.13            & 80.96           \\ & CoT                           & 72.54             & 77.97             & 77.45            & 81.53           \\
                              & \textbf{CogChain}                         & \textbf{75.98}    & \textbf{80.21}    & \textbf{80.34}   & \textbf{84.23}  \\ \midrule
\multirow{3}{*}{GPT-4o}    & Prompt                           & 73.21             & 77.79             & 78.86            & 81.29           \\   & CoT                           & 73.42             & 78.23             & 79.04            & 81.85           \\
                              & \textbf{CogChain}                         & \textbf{75.59}    & \textbf{80.35}    & \textbf{82.18}   & \textbf{84.90}  \\ \bottomrule
\end{tabular}
\caption{Performance of Our Cognition Chain. ``Prompt'' refers to instructing the LLM with direct prompt. ``CoT'' denotes prompting the LLM with the standard Chain-of-Thought. ``CogChain'' involves guiding the LLM to follow our Cognition Chain to derive the answer.}
\label{cogchain}
\end{table}

\begin{table}[]
\small
\begin{tabular}{@{}llllcccc@{}}
\toprule
\multirow{2}{*}{$\mathcal{S}$} & \multirow{2}{*}{$\mathcal{E}$} & \multirow{2}{*}{$\mathcal{R}$} & \multirow{2}{*}{$\mathcal{A}$} & \multicolumn{2}{c}{\textbf{Dreaddit}} & \multicolumn{2}{c}{\textbf{WBSD}} \\ \cmidrule(l){5-8} 
                   &                    &                    &                    & Acc         & F1-score           & Acc       & F1-score        \\ \midrule
\checkmark          & \checkmark          & \checkmark          & \checkmark          & \textbf{75.59}    & \textbf{80.35}    & \textbf{82.18}  & \textbf{84.90}  \\
\checkmark          & \checkmark          & $\times$           & \checkmark          & 74.36             & 79.24             & 80.37           & 83.51           \\
\checkmark          & $\times$           & \checkmark          & \checkmark          & 74.63             & 79.52             & 81.49           & 84.27           \\
\checkmark          & $\times$           & $\times$           & \checkmark          & 73.51             & 78.90             & 79.83           & 82.13           \\ 
$\times$          & $\times$           & $\times$           & \checkmark          & 72.70             & 77.54             & 78.67           & 81.15           \\ \bottomrule
\end{tabular}
\caption{Ablation Study. $\mathcal{S}$, $\mathcal{E}$, $\mathcal{R}$, and $\mathcal{A}$ denote the Stimulus, Evaluation, Reaction, and stress state steps of our Cognition Chain, respectively.}
\label{ablation}
\end{table}

\subsection{Effectiveness of Cognition Chain}
Table \ref{cogchain} presents the main results of our Cognition Chain approach. Prior research~\cite{wei2022chain} indicates that chain-of-thought only reasoning effects significantly when models exceed 100B parameters. Following this instruction, we selected three large-scale industry level LLMs as base models for comparison, which include GPT-3.5~\cite{gpt3.5}, Gemini-15~\cite{reid2024gemini}, and GPT-4o~\cite{gpt4o}. We compare our Cognition Chain method against both direct prompting~\cite{alghamdi2024less} and the standard chain-of-thought~\cite{wei2022chain} approach. Further details about the baselines are provided in Appendix \ref{app:baseline}.
Our results show that our Cognition Chain consistently achieves the highest performance in the comparison with all the baselines. Moreover, simply introducing the standard chain-of-thought into stress detection offers only marginal improvement over direct prompting, underscoring the value of our enhancements, which integrate Cognitive Appraisal Theory and construct a well-designed template to improve upon standard CoT reasoning.

We conduct further ablation study about the steps derived from the psychology routines within our Cognition Chain. The results in Table~\ref{ablation} show that including all four steps of the Cognition Chain (Stimulus, Evaluation, Reaction, and Stress State) consistently improves performance. When the full configuration is used, both accuracy and F1-scores are highest, indicating that each step contributes valuable information. As we remove steps, performance decreases steadily, confirming their complementary roles.  Notably, discarding the Evaluation and Reaction steps leads to a notable drop in scores, and relying solely on the Stress State step yields the lowest overall performance.
These findings highlight the importance of modelling all four steps to achieve accurate detection results.

\begin{table*}[]\centering
\small
\setlength\tabcolsep{8pt}
\begin{tabular}{lcccccccc}
\toprule
\multirow{2}{*}{\textbf{Model}} & \multicolumn{4}{c}{\textbf{Dreddit}}                                             & \multicolumn{4}{c}{\textbf{WBSD}}                                                 \\ \cmidrule(l){2-9} 
                                 & Accuracy & Precision & \multicolumn{1}{l}{Recall} & \multicolumn{1}{l}{F1-score} & Accuracy & Precision & \multicolumn{1}{l}{Recall} & \multicolumn{1}{l}{F1-score} \\ \midrule
\multicolumn{9}{c}{Proprietary LLMs} \\ \midrule
GPT-3.5                          & 73.25    & 71.34     & 85.54                      & 79.84                       & 79.25    & 76.22     & 89.13                      & 83.48                        \\
Gemini-1.5                       & 75.98    & 72.93     & 86.18                      & 80.21                        & 80.34    & 78.43     & 88.68                      & 84.23                        \\
GPT-4o                           & 75.59    & 73.60     & 87.52                      & 80.35                        & 82.18    & 77.16     & 91.37                      & 84.90                        \\ \midrule
\multicolumn{9}{c}{Open-source LLMs} \\ \midrule
Llama-2-7b                     & 69.51    & 69.53     & 79.61                      & 73.80                        & 75.32    & 75.64     & 79.17                      & 77.45                        \\
GLM-4-9b                     & 72.13    & 74.53     & 81.64                      & 77.31                        & 78.47    & 78.65     & 85.92                      & 82.06                        \\
Llama-3-8b                     & 74.16    & 74.09     & 82.83                      & 78.56                        & 80.38    & 78.01     & 86.63                      & 82.21                        \\ \midrule
\textbf{CogLLM (Ours)}           & \textbf{78.43} & \textbf{78.69} & \textbf{89.23}             & \textbf{83.15}               & \textbf{90.37} & \textbf{89.24} & \textbf{92.65}             & \textbf{90.61}               \\ \bottomrule
\end{tabular}
\caption{Performance of CogLLM. The first part lists proprietary LLMs. The second part list open-source LLMs. }
\label{tab:main}
\end{table*}

\subsection{Effectiveness of CogLLM}
Table \ref{tab:main} presents a summary of our CogLLM results. To conduct a comprehensive comparison, we employ two categories of baselines. The first category comprises proprietary LLMs including GPT-3.5~\cite{gpt3.5}, Gemini-1.5~\cite{reid2024gemini}, and GPT-4o~\cite{gpt4o}. The second category encompasses open-source LLMs such as Llama-2~\cite{touvron2023llama}, GLM-4~\cite{glm2024chatglm}, and Llama-3~\cite{dubey2024llama}. Details of the baselines are provided in Appendix \ref{app:baseline}.
The open-source LLMs are tuned via the same posts and labels as our CogLLM with the Cognition Chain data excluded. The results clearly indicate that our CogLLM significantly outperforms all compared models. When benchmarked against state-of-the-art proprietary LLMs such as GPT-4o, our model consistently attains higher accuracy and F1-scores, reflecting its effectiveness and adaptability. Furthermore, CogLLM establishes a new performance benchmark among open-source LLMs, surpassing models like Llama-3-8b by a considerable margin. Overall, these results underscore CogLLM’s strong capability for stress detection.

\begin{figure}[]
  \centering
\begin{subfigure}[t]{0.43\linewidth}
\centering
\includegraphics[width=1\textwidth]{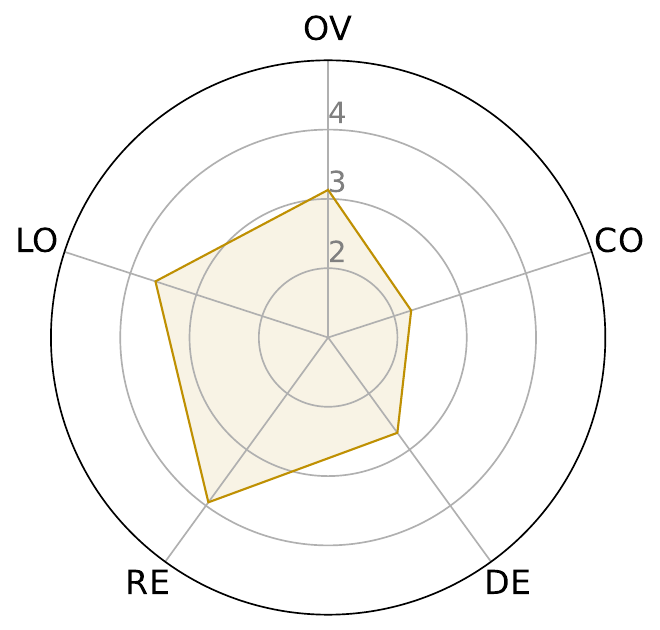}
\caption{LLama3}
\label{subfig:Llama3_radar}
\end{subfigure}
\begin{subfigure}[t]{0.43\linewidth}
\centering
\includegraphics[width=1\textwidth]{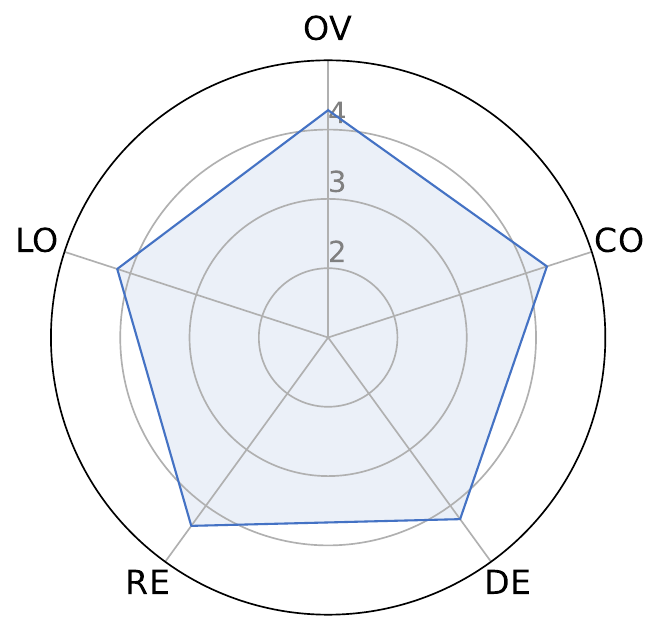}
\caption{CogLLM}
\end{subfigure}
  \caption{Human evaluation for explanations. The CO, DE, RE, LO, OV represent comprehension, depth, relevance, logic, overall, respectively. }
  \label{fig:human}
\end{figure}

\subsection{Evaluation for Explanations}
To assess the quality of the generated explanations, we consider five aspects for human evaluation~\cite{zhou2024measuring,cai2023improving,stenning2011reasoning}:
(\romannumeral1) \textit{Comprehension}. Comprehensive reasoning explanations for the generation of stress. (\romannumeral2) \textit{Depth}. Insight and reasonable inference beyond the literal text. (\romannumeral3) \textit{Relevance}. The relevance of the reasoning chains to the original post. (\romannumeral4) \textit{Logic}. The logical coherence within the reasoning chains.
(\romannumeral5) \textit{Overall}. The average evaluation of the above four aspects. 
Three professional evaluaters are asked to assign a score from \{1, 2, 3, 4, 5\} to 50 randomly sampled samples. 
For more details, please refer to Appendix \ref{appdenix: evaluater}.

As shown in Figure \ref{fig:human}, We select Llama-3-8b instructed by standard chain-of-thought as our baseline.
Across all evaluated criteria, CogLLM demonstrates consistently higher performance metrics than Llama-3-8b. Although Llama-3-8b’s performance in terms of Relevance and Logic is relatively close to that of CogLLM, it lags significantly in Comprehension and Depth. These results suggest that, due to our proposed CogInstruct and Cognition Chain, CogLLM can provide more comprehensive and insightful explanations for stress detection.

\section{Conclusion}
In this paper, we introduced the Cognition Chain, a method inspired by cognitive appraisal theory explaining the generation of stress step-by-step. Based on this, we further presented a three-stage self-reflective annotation pipeline to create CogInstruct, a new dataset for instruction-tuning large language models (LLMs), and developed a specialized model termed CogLLM. Experimental results validate the effectiveness of both the Cognition Chain and CogLLM in the comparison with the industry-level baselines. Our work marks an initial step toward leveraging cognitive theory for more explainable psychological detection, setting a foundation for future research to explore the intersection of LLM reasoning processes and cognitive theory. Ultimately, we hope this line of inquiry will foster more trustable and reliable approaches to modeling psychological phenomena.

\section{Limitation}
The quality and quantity of the cognition chain dataset are critical factors in determining the performance of our model. To minimize the cost associated with generating each cognition chain through human expertise, we have developed a three-stage self-reflection pipeline utilizing GPT-4o. A precise and lightweight classifier plays a important role in this pipeline; however, training such a classifier still necessitates a small amount of human effort for labelling. We believe that replacing the current classifier with a more advanced one may further improve data quality, though this improvement might require additional human labelling. Addressing these challenges remains a future objective, as we aim to develop a framework capable of generating high-quality datasets without human involvement.
 
\section{Ethics Statement}
In this study, all data utilized are sourced from publicly available social media platforms. The training and fine-tuning of large language models (LLMs) require substantial computational resources, which result in a significant environmental footprint. Increased energy consumption, along with the associated carbon emissions, represents an important ethical consideration in our work. To mitigate this concern, we have employed resource-efficient tuning strategies, such as LoRA (Low-Rank Adaptation), to minimize computational overhead wherever possible. We also acknowledge that while LLMs can generate informative and context-rich responses, they are not substitutes for licensed professionals in mental health or medicine. Specifically, we caution against using our tuned models as replacements for trained psychological practitioners. Any information or support provided by our LLMs should be considered an assistance to professional advice, not a replacement for it.
\section*{Acknowledgments}
This paper was supported by the Pandemic Sciences Institute at the University of Oxford; the National Institute for Health Research (NIHR) Oxford Biomedical Research Centre (BRC); an NIHR Research Professorship; a Royal Academy of Engineering Research Chair; the Wellcome Trust funded VITAL project (grant 204904/Z/16/Z); the EPSRC (grant EP/W031744/1); and the InnoHK Hong Kong Centre for Cerebro-cardiovascular Engineering (COCHE). 

\bibliography{CogChain}

\clearpage
\appendix

\newpage

\begin{table}[]
\small
\setlength\tabcolsep{3pt}
\begin{tabular}{@{}lcccc@{}}
\toprule
\multirow{2}{*}{\textbf{Model}} & \multicolumn{2}{c}{\textbf{Dreaddit}} & \multicolumn{2}{c}{\textbf{WBSD}} \\ \cmidrule(l){2-5} 
                                & Accuracy          & F1-score          & Accuracy        & F1-score        \\ \midrule
\textbf{CogLLM}                 & \textbf{78.43}    & \textbf{83.15}    & \textbf{90.37}  & \textbf{90.61}  \\ \midrule
w/o Human Filter                & 76.81             & 81.54             & 88.29           & 88.63           \\ \bottomrule
\end{tabular}
\caption{Ablation Study. ``w/o Human Filter'' indicates the removal of the human filter process during the construction of CogInstruct.} 
\label{tab:ablation_filter}
\end{table}
\section{Experiment about without Human Filter Process for CogInstruct}
\label{appdenix: filter}
To verify the effectiveness of the Human Filter Process in building the CogInstruct dataset, we conduct an experiment by instruction-tuning a model with the initial automatically generated data. As shown in Table \ref{tab:ablation_filter}, the performance drops without the human filter process. This verifies that the human filter process, assisted by the quality classifier, ensures data quality.

\section{Details of Human Evaluation}
\label{appdenix: evaluater}
We hired three master's students majoring in psychology to evaluate the explanations. We randomly sampled 50 generated explanations from Llama3 and CogLLM for evaluation. For each sample, we paid them \$0.3. The evaluation process was conducted under the supervision of a psychology expert. We report the average scores provided by the three evaluators.

\section{Baselines}
\label{app:baseline}
We compare our Cognition Chain with the following methods:
\begin{itemize}
    \item Direct Prompt \cite{alghamdi2024less}.This method uses a direct prompt to guide the LLM in generating stress detection results. The prompt is as follows:"\textit{Given the following social media text ( can be either post or post summary or post TLDR ), looking from the  poster's  perspective,  only  classify  if  it  is  indicative  of  current  very  severe  negative  stress  as  ‘Yes’  otherwise ‘No’. Just answer in ‘Yes’ or ‘No’. Don't provide explanations. Text: }"
    \item Standard COT \cite{wei2022chain}. This is the standard Chain-of-Thought (CoT) method. Its prompting consists of an instruction followed by a few examples with reasoning steps. We implement it as follows:
    \textit{I will input the individual's expression to you, please infer whether the individual is stressed or non-stressed. Let's think step by step. Please output "stressed" or "non-stressed". 
    [Examples]:...
Individual Expression: }
\end{itemize}
We compare our CogLLM with the following well-developed large language models:
\begin{itemize}
    \item GPT-3.5~\cite{gpt3.5}: This large, closed-source model is available through an API provided by OpenAI. We chose gpt-3.5-turbo, one of the most capable and cost-effective models in the GPT-3.5 family.
    \item Gemini-1.5~\cite{reid2024gemini}: This is the latest closed-source model published and available through the Google API. We picked the gemini-1.5-pro, the most widely used and commonly adopted one.
    \item GPT-4o~\cite{gpt4o}: This is OpenAI's flagship high-intelligence model, available through the OpenAI API. The version we picked is named gpt-4o.
    \item Llama-2~\cite{touvron2023llama}: This is a widely used open-source large language model developed by Meta. For this work, we selected the llama-2-7b version.
    \item GLM-4~\cite{glm2024chatglm}: This is a recently released flagship open-source large model from Zhipu AI. In this work, we used the glm-4-9b version.
    \item Llama-3~\cite{dubey2024llama}: This is a recently released flagship open-source large model by Meta. For our work, we selected the LLaMA3-8b-Instruct version.
\end{itemize}

\onecolumn
{\small
\phantomsection
\label{appx:Cognition Chain}
\begin{tcolorbox}[colback=white!95!gray,colframe=gray!50!black,rounded corners,label={prompt-Cognition Chain}, title={Prompt Template for Cognition Chain.}]
\begin{lstlisting}[breaklines=true, xleftmargin=0pt, breakindent=0pt, columns=fullflexible, mathescape]
Now you are a psychology expert.

I will input the individual's expression to you, please infer whether the individual is stressed. Note that you should understand and explain the process of stress generation step by step.

Specifically, Please perform a 4-step reasoning process:

1. Stimulus. Stimulus can be an event, object... If you indeed can not find a stimulus in the expression, you should output the Stimulus as 'N/A'.
2. Evaluation. There are three possible results for the individual's evaluation of the stimulus: "beneficial", "harmful" or "irrelevant" to the individual. If it is "beneficial", it will cause a positive emotional experience; if it is "harmful", it will cause a negative emotional experience; if it is "irrelevant", it will be ignored. 
3. Reaction. The individual's reaction or state and its emotions caused by the stimulus and evaluation. 
4. Stress state. Infer whether the individual is stressed based on the individual's reaction. The result should be "stressed" or "non-stressed".

I will give you some examples below:
----- Example -----
...

----- To be solved ----- 
Individual Expression: 

\end{lstlisting}
\end{tcolorbox}
}

{\small
\phantomsection
\label{appx:Self-Reflection}
\begin{tcolorbox}[colback=white!95!gray,colframe=gray!50!black,rounded corners,label={prompt-Self-Reflection}, title={Prompt Template for Self-Reflection}]
\begin{lstlisting}[breaklines=true, xleftmargin=0pt, breakindent=0pt, columns=fullflexible, mathescape]
Now you are a psychology expert.

I will input an individual's expression and the corresponding 4-step reasoning process to infer whether the individual is stressed. There exist some error in the steps. Please reflect on each step and then generate a new 4-step reasoning process for the expression.

Specifically, Please perform a 4-step reasoning process:

1. Stimulus. Stimulus can be an event, object... If you indeed can not find a stimulus in the expression, you should output the Stimulus as 'N/A'.
2. Evaluation. There are three possible results for the individual's evaluation of the stimulus: "beneficial", "harmful" or "irrelevant" to the individual. If it is "beneficial", it will cause a positive emotional experience; if it is "harmful", it will cause a negative emotional experience; if it is "irrelevant", it will be ignored. 
3. Reaction. The individual's reaction or state and its emotions caused by the stimulus and evaluation. 
4. Stress state. Infer whether the individual is stressed based on the individual's reaction. The result should be "stressed" or "non-stressed".

Please output the new process directly.

----- To be solved -----
Individual Expression: 

\end{lstlisting}
\end{tcolorbox}
}

{\small
\phantomsection
\label{appx:Answer-Reflection}
\begin{tcolorbox}[colback=white!95!gray,colframe=gray!50!black,rounded corners,label={prompt-Answer-Reflection}, title={Prompt Template for Answer-Reflection}]
\begin{lstlisting}[breaklines=true, xleftmargin=0pt, breakindent=0pt, columns=fullflexible, mathescape]
Now you are a psychology expert.

I will input an individual's expression, the corresponding 4-step reasoning process used to infer whether the individual is stressed, and the real stress state. The provided reasoning process is incorrect. Please reflect on the errors in it and then generate a correct 4-step reasoning process based on the real stress state.

Specifically, Please perform a 4-step reasoning process:

1. Stimulus. Stimulus can be an event, object... If you indeed can not find a stimulus in the expression, you should output the Stimulus as 'N/A'.
2. Evaluation. There are three possible results for the individual's evaluation of the stimulus: "beneficial", "harmful" or "irrelevant" to the individual. If it is "beneficial", it will cause a positive emotional experience; if it is "harmful", it will cause a negative emotional experience; if it is "irrelevant", it will be ignored. 
3. Reaction. The individual's reaction or state and its emotions caused by the stimulus and evaluation. 
4. Stress state. Infer whether the individual is stressed based on the individual's reaction. The result should be "stressed" or "non-stressed".

Please output the new process directly.

----- To be solved -----
Individual Expression: 

\end{lstlisting}
\end{tcolorbox}
}
\clearpage
\begin{figure*}[h]
\centering 
\includegraphics[width=1\linewidth]{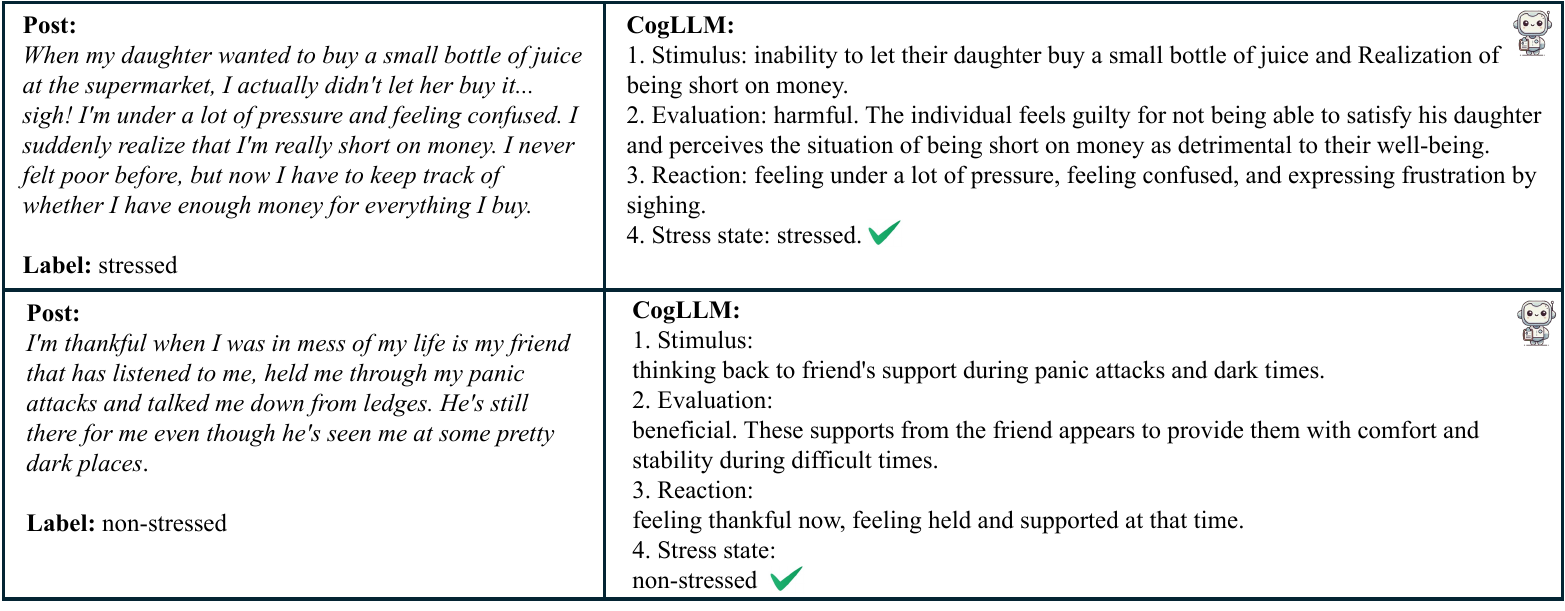}
\caption{Case Study of CogLLM}
\label{fig:casestudy}
\end{figure*}


\end{document}